\documentstyle[aclap]{article}

\begin{document}

\title{{\bf A MATCHING TECHNIQUE IN EXAMPLE-BASED MACHINE TRANSLATION}}

\author{Lambros CRANIAS, Harris PAPAGEORGIOU, Stelios PIPERIDIS \\
Institute for Language and Speech Processing, GREECE\\
Stelios.Piperidis@eurokom.ie}
\date{}

\maketitle

\section*{ABSTRACT}

This paper addresses an important problem in Example-Based Machine Translation
(EBMT), namely how to measure similarity between a sentence fragment and a set
of stored examples. A new method is proposed that measures similarity according
to both surface structure and content. A second contribution is the use of
clustering to make retrieval of the best matching example from the database
more efficient. Results on a large number of test cases from the CELEX database
are presented.

\section{INTRODUCTION}

EBMT is based on the idea of performing translation by imitating translation
examples of similar sentences [Nagao 84]. In this type of translation system, a
large amount of bi/multi-lingual translation examples has been stored in a
textual database and input expressions are rendered in the target language by
retrieving from the database that example which is most similar to the input.

There are three key issues which pertain to example-based translation:

\begin{itemize}
\item establishment of correspondence between units in a bi/multi-lingual text
at sentence, phrase or word level
\item  a mechanism for retrieving from the database the unit that best matches
the input
\item exploit the retrieved translation example to produce the actual
translation of the input sentence
\end{itemize}
[Brown 91] and [Gale 91] have proposed methods for establishing correspondence
between sentences in bilingual corpora. [Brown 93], [Sadler 90] and [Kaji 92]
have tackled the problem of establishing correspondences between words and
phrases in bilingual texts.

The third key issue of EBMT, that is exploiting the retrieved translation
example, is usually dealt with by integrating into the system conventional MT
techniques [Kaji 92], [Sumita 91]. Simple modifications of the translation
proposal, such as word substitution, would also be possible, provided that
alignment of the translation archive at word level was available.

In establishing a mechanism for the best match retrieval, which is the topic of
this paper, the crucial tasks are: (i) determining whether the search is for
matches at sentence or sub-sentence level, that is determining the ``text
unit'', and (ii) the definition of the metric of similarity between two text
units.

As far as (i) is concerned, the obvious choice is to use as text unit the
sentence. This is because, not only are sentence boundaries unambiguous but
also translation proposals at sentence level is what a translator is usually
looking for. Sentences can, however, be quite long. And the longer they are,
the less possible it is that they will have an exact match in the translation
archive, and the less flexible the EBMT system will be.

On the other hand if the text unit is the sub-sentence we face one major
problem, that is the possibility that the resulting translation of the whole
sentence will be of low quality, due to boundary friction and incorrect
chunking. In practice, EBMT systems that operate at sub-sentence level involve
the dynamic derivation of the optimum length of segments of the input sentence
by analysing the available parallel corpora. This requires a procedure for
determining the best ``cover'' of an input text by
segments of sentences contained in the database [Nirenburg 93]. It is assumed
that the translation of the segments of the database that cover the input
sentence is known. What is needed, therefore, is a procedure for aligning
parallel texts at sub-sentence level [Kaji 92], [Sadler 90]. If sub-sentence
alignment is available, the approach is fully automated but is quite vulnerable
to the problem of low quality as mentioned above, as well as to ambiguity
problems when the produced segments are rather small.

Despite the fact that almost all running EBMT systems employ the sentence as
the text unit, it is believed that the potential of EBMT lies on the
exploitation of fragments of text smaller that sentences and the combination of
such fragments to produce the translation of whole sentences [Sato 90].
Automatic sub-sentential alignment is, however, a problem yet to be solved.

Turning to the definition of the metric of similarity, the requirement is
usually twofold. The similarity metric applied to two sentences (by sentence
from now on we will refer to both sentence and sub-sentence fragment) should
indicate how similar the compared sentences are, and perhaps the parts of the
two sentences that contributed to the similarity score. The latter could be
just a useful indication to the translator using the EBMT system, or a crucial
functional factor of the system as will be later
explained.

The similarity metrics reported in the literature can be characterised
depending on the text patterns they are applied on. So, the word-based metrics
compare individual words of the two sentences in terms of their morphological
paradigms, synonyms, hyperonyms, hyponyms, antonyms, pos tags... [Nirenburg 93]
or use a semantic distance d (0 $\leq$ d $\leq$ 1) which is determined by the
Most Specific Common Abstraction (MSCA) obtained from a thesaurus abstraction
hierarchy
[Sumita 91]. Then, a similarity metric is devised, which reflects the
similarity of two sentences, by combining the individual contributions towards
similarity stemming from word comparisons.

The word-based metrics are the most popular, but other approaches include
syntax-rule driven metrics [Sumita 88], character-based metrics [Sato 92] as
well as some hybrids [Furuse 92]. The character-based metric has been applied
to Japanese, taking advantage of certain characteristics of the Japanese. The
syntax-rule driven metrics try to capture similarity of two sentences at the
syntax level. This seems very promising, since similarity at the syntax level,
perhaps coupled by lexical similarity in a
hybrid configuration, would be the best the EBMT system could offer as a
translation proposal. The real time feasibility of such a system is, however,
questionable since it involves the complex task of syntactic analysis.

In section 2 a similarity metric is proposed and analysed. The statistical
system presented consists of two phases, the Learning and the decision making
or Recognition phase, which are described in section III. Finally, in section
IV the experiment configuration is discussed and the results evaluated.

\section{THE SIMILARITY METRIC}

To encode a sentence into a vector, we exploit information about the functional
words/phrases (fws) appearing in it, as well as about the lemmas and pos
(part-of-speech) tags of the words appearing between fws/phrases. Based on the
combination of fws/phrases data and pos tags, a simple view of the surface
syntactic structure of each sentence is obtained.

To identify the fws/phrases in a given corpus the following criteria are
applied:

\begin{itemize}
\item fws introduce a syntactically standard behaviour
\item most of the fws belong to closed classes
\item the semantic behaviour of fws is determined through their context
\item most of the fws determine phrase boundaries
\item  fws have a relatively high frequency in the corpus
\end{itemize}

According to these criteria, prepositions, conjunctions, determiners, pronouns,
certain adverbials etc. are regarded as fws. Having identified the fws of the
corpus we distinguish groups of fws on the basis of their interchangeability in
certain phrase structures. The grouping caters, also, for the multiplicity of
usages of a certain word which has been identified as a fw, since a fw can be a
part of many different groups. In this way, fws can serve the retrieval
procedure with respect to the following
two levels of contribution towards the similarity score of two sentences :

\begin{itemize}
\item  Identity of fws of retrieved example and input (I)
\item fws of retrieved example and input not identical but belonging to the
same group (G)
\end{itemize}

To obtain the lemmas and pos tags of the remaining words in a sentence, we use
a part-of-speech Tagger with no disambiguation module, since this would be time
consuming and not 100\% accurate. Instead, we introduce the concept of
ambiguity class (ac) and we represent each non-fw by its ac and the
corresponding lemma(s) (for example, the unambiguous word ``eat'' would be
represented by the ac which is the set {verb} and the lemma ``eat'') (in
English, for an ambiguous word, the corresponding lemmas will
usually be identical. But this is rarely true for Greek). Hence, the following
two levels of contribution to the similarity score stem from non-fws:

\begin{itemize}
\item overlapping of the sets of possible lemmas of the two words (L)
\item overlapping of the ambiguity classes of the two words (T)
\end{itemize}

Hence, each sentence of the source part of the translation archive is
represented by a pattern, which is expressed as an ordered series of the above
mentioned feature components.

A similarity metric is defined between two such vectors, and is used in both
the Learning and Recognition phases. Comparing a test vector against a
reference vector is, however, not straightforward, since there are generally
axis fluctuations between the vectors (not necessarily aligned vectors and of
most probably different length). To overcome these problems we use a two-level
Dynamic Programming (DP) technique [Sakoe 78], [Ney 84]. The first level treats
the matches at fw level, while the second is
reached only in case of a match in the first level, and is concerned with the
lemmas and tags of the words within fw boundaries. Both levels utilise the same
(DP) model which is next described.

We have already referred to the (I) and (G) contributions to the similarity
score due to fws. But this is not enough. We should also take into account
whether the fws appear in the same order in the two sentences, whether an extra
(or a few) fws intervene in one of the two sentences, whether certain fws are
missing ... To deal with these problems, we introduce a yet third contribution
to the similarity score, which is negative and is called penalty score (P). So,
as we are moving along a diagonal of the
xy-plane (corresponding to matched fws), whenever a fw is mismatched, it
produces a negative contribution to the score along a horizontal or vertical
direction. In figure 1 the allowable transitions in the xy-plane are shown:

\begin{figure}[htb]
\setlength{\unitlength}{4mm}
\begin{picture}(14,6)(0,0)
\put(7,0){\vector(1,1){4}}
\put(12,0){\vector(0,1){4}}
\put(7,5){\vector(1,0){4}}
\put(7,0){\line(1,1){5}}
\put(12,4){\line(0,1){1}}
\put(11,5){\line(1,0){1}}
\put(10,6){P}
\put(13,3){P}
\put(7.6,3){I (G)}
\end{picture}
\caption {The DP allowable transitions}
\end{figure}
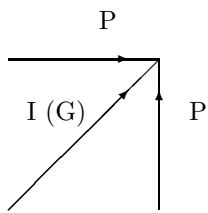

Whenever a diagonal transition is investigated, the system calls the second
level DP-algorithm which produces a local additional score due to the potential
similarity of lemmas and tags of the words lying between the corresponding fws.
This score is calculated using exactly the same DP-algorithm as the one
treating fws (allowing additions, deletions,...), provided that we use (L), (T)
and (PT) (a penalty score attributed to a mismatch at the tag-level) in place
of (I), (G) and  (P) respectively.

The outcome of the DP-algorithm is the similarity score between two vectors
which allows for different lengths of the two sentences, similarity of
different parts of the two sentences (last part of one with the first part of
the other) and finally variable number of additions and deletions. The score
produced, corresponds to two coherent parts of the two sentences under
comparison. Emphasis should be given to the variable number of additions and
deletions. The innovation of the penalty score (which is
in fact a negative score) provides the system with the flexibility to afford a
different number of additions or deletions depending on the accumulated
similarity score up to the point where these start. Moreover, the algorithm
determines, through a backtracking procedure, the relevant parts of the two
vectors that contributed to this score. This is essential for the sentence
segmentation described in the next section.

It should also be noted that the similarity score produced is based mainly on
the surface syntax of the two sentences (as this is indicated by the fws and
pos tags) and in the second place on the actual words of the two sentences.
This is quite reasonable, since the two sentences could have almost the same
words in the source language but no similarity at all in the source or target
language (due to different word order, as well as different word utilisation),
while if they are similar in terms of fws as
well as in terms of the pos tags of the words between fws, then the two
sentences would almost certainly be similar (irrelevant of a few differences in
the actual words) in the target language as well (which is the objective).

The DP-algorithm proposed seems to be tailored to the needs of the similarity
metric but there is yet a crucial set of parameters to be set, that is
A={I,G,P,L,T,PT}. The DP-algorithm is just the framework for the utilisation of
these parameters. The values of the parameters of A are set dynamically
depending on the lengths of the sentences under comparison. I, G, L, T are set
to values (I, G are normalised by the lengths of the sentences in fws, while L,
T are normalised by the lengths of the blocks of
words appearing between fws) which produce a 100\% similarity score when the
sentences are identical, while P, PT reflect the user's choise of penalising an
addition or deletion of a word (functional or not).

\section{LEARNING AND RECOGNITION PHASES}

In the Learning phase, the modified k-means clustering procedure [Wilpon 85] is
applied to the source part of the translation archive, aiming to produce
clusters of sentences, each represented by its centre only. The algorithm
produces the optimum segmentation of the corpus into clusters (based on the
similarity metric), and determines each cluster centre (which is just a
sentence of the corpus) by using the minmax criterion. The number of clusters
can be determined automatically by the process, subject
to some cluster quality constraint (for example, minimum intra-cluster
similarity), or alternatively can be determined externally based upon
memory-space restrictions and speed requirements.

Once the clustering procedure is terminated, a search is made, among the
sentences allocated to a cluster, to locate second best (but good enough)
matches to the sentences allocated to the remaining clusters. If such matches
are traced, the relevant sentences are segmented and then the updated corpus is
reclustered. After a number of iterations, convergence is obtained (no new
sentence segments are created) and the whole clustering procedure is
terminated.

Although the objective of a matching mechanism should be to identify in a
database the longest piece of text that best matches the input, the rationale
behind sentence segmentation is in this case self-evident. It is highly
probable that a sentence is allocated to a cluster center because of a good
match due to a part of it, while the remaining part has nothing to do with the
cluster to which it will be allocated. Hence, this part will remain hidden to
an input sentence applied to the system at the
recognition phase. On the other hand, it is also highly probable that a given
input sentence does not, as a whole, match a corpus sentence, but rather
different parts of it match with segments belonging to different sentences in
the corpus. Providing whole sentences as translation proposals, having a part
that matched with part of the input sentence, would perhaps puzzle the
translator instead of help him (her).

But sentence segmentation is not a straightforward matter. We can not just
segment a sentence at the limits of the part that led to the allocation of the
sentence to a specific cluster. This is because we need to know the translation
of this part as well. Hence, we should expand the limits of the match to cover
a ``translatable unit'' and then segment the sentence. Automatic sub-sentential
alignment (which would produce the ``translatable units''), however, is not yet
mature enough to produce high
fidelity results. Hence, one resorts to the use of semi-automatic methods (in
our application with the CELEX database, because of the certain format in which
the texts appear, a rough segmentation of the sentences is straightforward and
can therefore be automated).

If alignment at sub-sentential level is not available, the segmentation of the
sentences of the corpus is not possible (it is absolutely pointless). Then, the
degree of success of the Learning phase will depend on the length of the
sentences contained in the corpus. The longer these sentences tend to be, the
less successful the Learning phase. On the other hand, if alignment at
sub-sentential level is available, we could just apply the clustering procedure
to these segments. But then, we might end up
with an unnecessary large number of clusters and ``sentences''. This is
because, in a specific corpus quite a lot of these segments tend to appear
together. Hence, by clustering whole sentences and then segmenting only in case
of a good match with a part of  a sentence allocated to a different cluster, we
can avoid the overgeneration of clusters and segments. When the iterative
clustering procedure is finally terminated,  the sentences of the original
corpus will have been segmented to ``translatable
units'' in an optimum way, so that they are efficiently represented by a set of
sentences which are the cluster centres.

In the Recognition phase, the vector of the input sentence is extracted and
compared against the cluster centres. Once the favourite cluster(s) is
specified, the search space is limited to the sentences allocated to that
cluster only, and the same similarity metric is applied to produce the best
match available in the corpus. If the sentences in the translation archive have
been segmented, the problem is that, now, we do not know what the
``translatable units'' of the input sentence are (since we do not
know its target language equivalent). We only have potential ``translatable
unit'' markers. This is not really a restriction, however, since by setting a
high enough threshold for the match with a segment (translatable piece of text)
in the corpus, we can be sure that the part of the input sentence that
contributed to this good match, will also be translatable and we can,
therefore, segment this part. This process continues until the whole input
sentence has been ``covered'' by segments of the corpus.

\section{APPLICATION-EVALUATION}

The development of the matching method presented in this paper was part of the
research work conducted under the LRE I project TRANSLEARN. The project will
initially consider four languages: English, French, Greek and Portugese. The
application on which we are developing and testing the method is implemented on
the Greek-English language pair of records of the CELEX database, the
computerised documentation system on Community Law, which is available in all
Community languages. The matching mechanism is,
so far, implemented on the Greek part, providing English translation proposals
for Greek input sentences. The sentences contained in the CELEX database tend
to be quite long, but due to the certain format in which they appear
(corresponding to articles, regulations,...), we were able to provide the
Learning phase with some potential segmentation points of these sentences in
both languages of the pair (these segmentation points are in one-to-one
correspondence across languages, yielding the
``sub-sentence'' alignment).

In tagging the Greek part of the CELEX database we came across 31 different
ambiguity classes, which are utilised in the matching mechanism. The
identification and grouping of the Greek fws was mainly done with the help of
statistical tools applied to the CELEX database.

We tested the system on 8,000 sentences of the CELEX database. We are
presenting results on two versions. One of 80 clusters (which accounts for the
1\% of the number of the sentences of the corpus used) which resulted in 10,203
``sentences'' (sentences or segments) in 2 iterations, and one of 160 clusters
which resulted in 10,758 ``sentences'' in 2 iterations. To evaluate the system,
we asked five translators to assign each translation proposal of the system (in
our application these proposals sometimes
refer to segments of the input sentence) to one of four categories :

\noindent A : The proposal is the correct (or almost) translation\\
B : The proposal is very helpful in order to produce the translation\\
C : The proposal can help in order to produce the translation\\
D : The proposal is of no use to the translator\\

We used as test suite 200 sentences of the CELEX database which were not
included in the translation archive. The system proposed translations for 232
``sentences'' (segments or whole input sentences) in the former case and for
244 in the latter case. The results are tabulated in table 1 (these results
refer to the single best match located in the translation archive):

\begin{center}

Table 1

\begin{small}
\begin{tabular}{|l|l|l|} \hline
    &  {\bf 80 CLUSTERS} & {\bf 160 CLUSTERS} \\ \hline
{\bf A} & {\bf 220 (19\%)} & {\bf 244 (20\%)} \\ \hline
{\bf B} & {\bf 464 (40\%)} & {\bf 512 (42\%)} \\ \hline
{\bf C} & {\bf 209 (18\%)} & {\bf 245 (20\%)} \\ \hline
{\bf D} & {\bf 267 (23\%)} & {\bf 219 (18\%)} \\ \hline
  & {\bf 1160} & {\bf 1220} \\ \hline
\end{tabular}
\end{small}
\end{center}

The table shows that in the case of 160 clusters, (1) at 62\% the system will
be very useful to the translator, and (2) some information can at least be
obtained from 82\% of the retrievals. In the case of 80 clusters the results do
not change significantly. Hence, as far as the similarity metric is concerned
the results seem quite promising (it should, however, be mentioned, that the
CELEX database is quite suitable for EBMT applications, due to its great degree
of repetitiveness).

On the other hand, the use of clustering of the corpus dramatically decreases
the response time of the system, compared to the alternative of searching
exhaustively through the corpus. Other methods for limiting the search space do
exist (for example, using full-text retrieval based on content words), but are
rather lossy, while clustering provides an effective means of locating the best
available match in the corpus (in terms of the similarity metric employed).
This can be seen in Table 2, where the
column  ``MISSED'' indicates the percentage of the input ``sentences'' for
which the best match in the corpus was not located in the favourite cluster,
while the column ``MISSED BY'' indicates the average deviation of the located
best matches from the actual best matches in the corpus for these cases.

\begin{center}

Table 2

\begin{small}
\begin{tabular}{|l|l|l|} \hline
    & {\bf MISSED} & {\bf MISSED BY} \\ \hline
{\bf 80 CLUSTERS} & {\bf 10\%} & {\bf 6.32\%} \\ \hline
{\bf 160 CLUSTERS} & {\bf 8.5\%} & {\bf 6.14\%} \\ \hline
\end{tabular}
\end{small}
\end{center}

In Table 1 as well as in Table 2 it can be seen that a quite important decrease
in the number of clusters affected the results only slightly. This small
deterioration in the performance of the system is due to ``hidden'' parts of
sentences allocated to clusters (parts that are not represented by the cluster
centres). Hence, the smaller the ``sentences'' contained in the database and
the more the clusters, the better the performance of the proposed system. The
number of clusters, however, should be
constrained for the search space to be effectively limited.

\section{REFERENCES}

[BROWN 91] Brown P. F. et al, (1991). ``Aligning Sentences in Parallel
Corpora''. {\em Proc. of the 29th Annual Meeting of the ACL}, pp 169-176.

[BROWN 93] Brown P. F. et al, (June 1993). ``The mathematics of Statistical
Machine Translation: Parameter Estimation''. {\em Computational Linguistics},
pp 263-311.

[FURUSE 92] Furuse O. and H. Iida, (1992). ``Cooperation between Transfer and
Analysis in Example-Based Framework''. {\em Proc. Coling}, pp 645-651.

[GALE 91] Gale W. A. and K. W. Church, (1991). ``A Program for Aligning
Sentences in Bilingual Corpora''. {\em Proc. of the 29th Annual Meeting of the
ACL.}, pp 177-184.

[KAJI 92] Kaji H., Y. Kida and Y. Morimoto, (1992). ``Learning Translation
Templates from Bilingual Text''. {\em Proc. Coling.}, pp 672-678.

[NAGAO 84] Nagao M., (1984). ``A framework of a mechanical translation between
Japanese and English by analogy principle''. {\em Artificial and Human
Intelligence}, ed. Elithorn A. and Banerji R., North-Holland, pp 173-180.

[NEY 84] Ney H., (1984). ``The use of a One-stage Dynamic Programming Algorithm
for Connected Word Recognition''. IEEE vol. ASSP-32, No 2.

[NIRENBURG 93] Nirenburg S. et al, (1993). ``Two Approaches to Matching in
Example-Based Machine Translation''. {\em Proc. of TMI-93, Kyoto, Japan}.

[SADLER 90] Sadler V. and R. Vendelmans, (1990). ``Pilot Implementation of a
Bilingual Knowledge Bank''. {\em Proc. of Coling}, pp 449-451.

[SAKOE 78] Sakoe H. and S. Chiba, (1978). ``Dynamic Programming Algorithm
Optimisation for Spoken Word Recognition''. {\em IEEE Trans. on ASSP, vol.
ASSP-26}.

[SATO 90] Sato S. and M. Nagao, (1990). ``Toward Memory-based Translation''.
{\em Proc. of Coling}, pp 247-252.

[SATO 92] Sato S., (1992). ``CTM: An Example-Based Translation Aid System''.
{\em Proc. of Coling}, pp 1259-1263.

[SUMITA 88] Sumita E. and Y. Tsutsumi, (1988). ``A Translation Aid System Using
Flexible Text Retrieval Based on Syntax-Matching''. {\em TRL Research Report},
Tokyo Research Laboratory, IBM.

[SUMITA 91] Sumita E. and H. Iida, (1991). ``Experiments and Prospects of
Example-based Machine Translation''. {\em Proc. of the 29th Annual Meeting of
the Association for Computational Linguistics}, pp 185-192.

[WILPON 85] Wilpon J. and L. Rabiner, (1985). ``A Modified k-Means Clustering
Algorithm for Use in Isolated Word Recognition''. {\em IEEE vol. ASSP-33}, pp.
587-594.

\end{document}